\documentclass[10pt,twocolumn,letterpaper]{article}

\usepackage[pagenumbers]{cvpr} 

\usepackage{capt-of}
\usepackage[dvipsnames]{xcolor}
\usepackage{xspace}
\usepackage{enumitem}
\usepackage{amsmath,amssymb,amsbsy,amsfonts,dsfont,pifont,bm,bbm,mathrsfs,mathtools,nicefrac}
\usepackage{algorithm,algpseudocode,listings}
\usepackage{booktabs,multirow,adjustbox,diagbox,threeparttable,tabularray}
\usepackage{bbding}
\usepackage{enumitem}

\newcommand{\tocite}[1]{{\color{red} [TO CITE]}}

\newcommand{\methodname}{DepthLab}
\newcommand{\method}{\texttt{\methodname}\xspace}

\definecolor{cvprblue}{rgb}{0.21,0.49,0.74}
\usepackage[pagebackref,breaklinks,colorlinks,citecolor=cvprblue]{hyperref}
\usepackage{wrapfig}
\usepackage[capitalize]{cleveref}  
\crefname{section}{Sec.}{Secs.}
\Crefname{section}{Section}{Sections}
\crefname{table}{Tab.}{Tabs.}
\Crefname{table}{Table}{Tables}
\crefname{figure}{Fig.}{Figs.}
\Crefname{figure}{Figure}{Figures}
\crefname{equation}{Eq.}{Eqs.}
\Crefname{equation}{Equation}{Equations}
\title{\methodname: From Partial to Complete}

\author{
Zhiheng Liu$^{1*}$, Ka Leong Cheng$^{2,3*}$, Qiuyu Wang$^{2}$, Shuzhe Wang$^{4}$, Hao Ouyang$^{2}$,\\
Bin Tan$^2$, Kai Zhu$^{5}$, Yujun Shen$^{2}$, Qifeng Chen$^{3\dag}$, Ping Luo$^{1\dag}$\\
\\
$^1$HKU, $^2$Ant Group, $^3$HKUST, $^4$Aalto University, $^5$Tongyi Lab\\
}

\begin{document}

\twocolumn[{
\renewcommand\twocolumn[1][]{#1}
\maketitle
\begin{center}
    \vspace{-13pt}
    \includegraphics[width=0.99\linewidth]{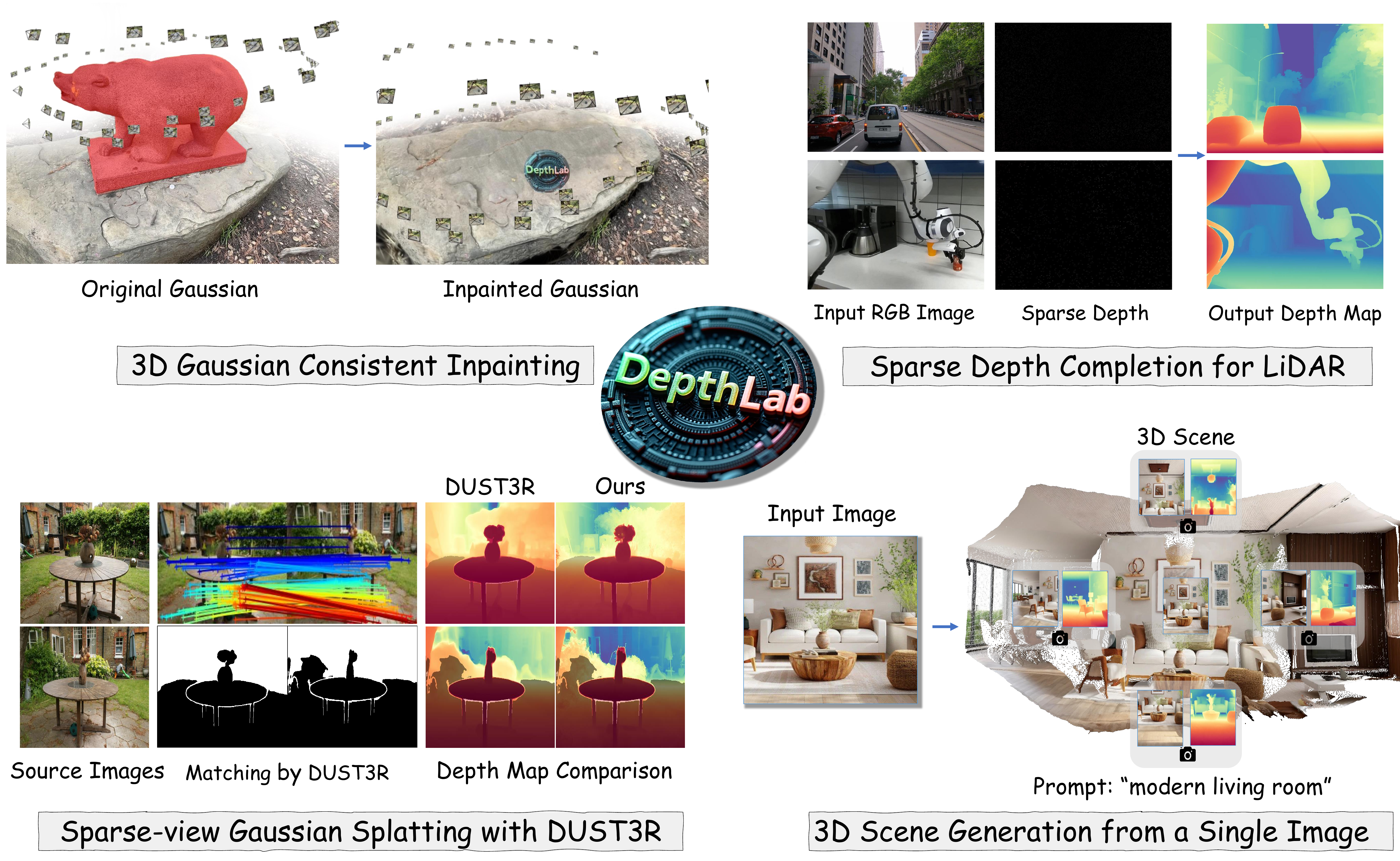}
    \captionsetup{type=figure}
    \vspace{-8pt}
    \caption{\textbf{\method~ for diverse downstream tasks.} Many tasks naturally contain partial depth information, such as (1) 3D Gaussian inpainting, (2) LiDAR depth completion, (3) sparse-view reconstruction with Dust3R, and (4) text-to-scene generation. 
    Our model leverages this known information to achieve improved depth estimation, enhancing performance in downstream tasks. We hope to motivate more related tasks to adopt \method. 
    }
    \label{fig:teaser}
    \vspace{8pt}
\end{center}
}]

\begin{abstract}

Missing values remain a common challenge for depth data across its wide range of applications, stemming from various causes like incomplete data acquisition and perspective alteration.
This work bridges this gap with \method, a foundation depth inpainting model powered by image diffusion priors.
Our model features two notable strengths:
(1) it demonstrates resilience to depth-deficient regions, providing reliable completion for both continuous areas and isolated points,
and (2) it faithfully preserves scale consistency with the conditioned known depth when filling in missing values.
Drawing on these advantages, our approach proves its worth in various downstream tasks, including 3D scene inpainting, text-to-3D scene generation, sparse-view reconstruction with DUST3R, and LiDAR depth completion, exceeding current solutions in both numerical performance and visual quality.
Our project page with source code is available at~\url{https://johanan528.github.io/depthlab_web/}.

\end{abstract}

\section{Introduction}
%
%
%
Depth inpainting, the task of reconstructing missing or occluded depth information in images, is pivotal in numerous domains, including 3D vision~\cite{li2019high,izadi2011kinectfusion,zhou2014color}, robotics~\cite{bagnell2010learning, kim2015deep}, and augmented reality~\cite{slater1997framework,rasla2022relative}.
%
%
As shown in~\cref{fig:teaser}, a robust depth inpainting model can enable high-quality 3D scene completion, editing, reconstruction, and generation. 
Previous research in depth inpainting can be categorized into two primary approaches.
The first approach~\cite{tang2024bilateral,wang2024improving,zhang2023completionformer, park2020non, de2021deep, jeon2022struct} focuses on completing globally sparse LiDAR depth~\cite{geiger2012we} data to dense depth, typically trained and tested on fixed datasets. 
However, these models lack generalization and have limited applicability across diverse downstream tasks. 
The second approach~\cite{marigold,xu2024diffusion,geowizard,depthanything,yang2024depthv2} employs monocular depth estimators to infer the depth of a single image, aligning the inpainted region with known depth.
These methods often suffer from significant geometric inconsistencies, particularly along edges, due to misalignment between estimated depth and existing geometry. 
Recent research~\cite{infusion} incorporates RGB images into the U-Net input as guidance to train a depth inpainting model, but its performance remains suboptimal in complex scenes and when inpainting large missing regions.
To this end, we propose \textbf{\method}, a foundation model for RGB image-conditioned depth inpainting, which introduces a dual-branch depth inpainting diffusion framework.
Specifically, the framework processes a single reference image through a Reference U-Net, which extracts RGB features as conditional input. Simultaneously, the known depth and the mask-indicating regions requiring inpainting are fed into the depth Estimation U-Net. The extracted RGB features are progressively integrated layer by layer into the depth Estimation U-Net to guide the inpainting process
During training, we apply random scale normalization to the known depth to mitigate regularization overflow caused by non-global extrema in the known regions.
Similar to Marigold~\cite{marigold}, our model requires only synthetic RGB-D data for training with just a few GPU days. Benefiting from the powerful priors of the diffusion model, \method~demonstrates strong generalization capabilities across diverse scenarios.

As shown in~\cref{fig:teaser}, thanks to accurate depth inpainting, \method~supports a variety of downstream applications.
%
%
(1) \textit{3D scene inpainting}~\cite{infusion}: In 3D scenes, we start by inpainting the depth of the image inpainted regions from the posed reference views, then unproject the points into the 3D space for optimal initialization, which significantly enhances the quality and speed of the 3D scene inpainting.
(2) \textit{Text-to-scene generation}~\cite{yu2023wonderjourney}: Our method substantially improves the process of generating a 3D scene from a single image by eliminating the need for alignment. This advancement effectively mitigates issues of edge disjunction that previously arose from geometric inconsistencies between the inpainted and known depth, thereby significantly enhancing the coherence and quality of the generated scenes.
(3) \textit{Sparse-View Reconstruction with DUST3R}:
%
InstantSplat~\cite{fan2024instantsplat} leverages point clouds from DUST3R~\cite{dust3r_cvpr24} as an initialization for SfM-free reconstruction and novel view synthesis. 
By adding noise to DUST3R depth maps as a latent input, our method refines depth in regions lacking cross-view correspondences, producing more precise, geometry-consistent depth maps. 
These refined depth maps further enhance the initial point clouds for InstantSplat.
(4) \textit{LiDAR depth completion}: Sensor depth completion~\cite{zhang2023completionformer} is an important task related to depth estimation. Unlike existing methods that are trained and tested on a single dataset, such as NYUv2~\cite{Silberman:ECCV12}, our approach achieves comparable results in a zero-shot setting and can deliver even better outcomes with minimal fine-tuning.
For a more comprehensive evaluation, we assessed its effectiveness on depth estimation benchmarks by applying various types of random masks to regions requiring inpainting.

%

%

\section{Related Work}
\subsection{Depth Completion}
Depth completion~\cite{ma2019self, ma2018sparse, hu2021penet, liu2021fcfr, wang2023lrru, yan2024tri} is a key task in computer vision, especially with the rise of active depth sensors. It aims to estimate accurate depth maps from sparse ground-truth measurements provided by sensors. However, existing methods~\cite{park2020non, de2021deep,jeon2022struct,zhao2021adaptive,zhang2023completionformer,tang2024bilateral,wang2023lrru} are often trained and evaluated on fixed datasets, limiting their generalization abilities, and many of them rely on cumbersome processing pipelines. In contrast, this paper offers an efficient and robust foundational model for depth completion, capable of strong generalization across diverse scenarios in various downstream applications.

\subsection{Monocular Depth Estimation}
Monocular depth estimation is a crucial task in computer vision, approached mainly through discriminative and generative methods.
Discriminative methods often predict depth as either metric depth~\cite{bhat2021adabins, bhat2022localbins, jun2022depth,li2024binsformer,mertan2022single,yuan2022new} or relative depth~\cite{yin2020diversedepth,ranftl2020towards,yin2021learning,eftekhar2021omnidata,zhang2022hierarchical,ranftl2021vision,yang2024depth,yang2024depthv2}. 
%
Recently, generative models~\cite{bochkovskii2024depth, hu2024depthcrafter, zhang2024betterdepth}, particularly diffusion models, gained popularity in depth estimation. Methods like DDP~\cite{ji2023ddp}, DepthGen~\cite{saxena2023monocular}, Marigold~\cite{marigold}, and GeoWizard~\cite{geowizard} leverage diffusion processes to produce high-quality depth maps, although often at the cost of slow sampling. Flow matching-based models~\cite{gui2024depthfm} emerge to mitigate these limitations, providing faster sampling.
%
%
Despite advances in monocular depth estimation, aligning estimated depth with known data often causes geometric inconsistency. Our work addresses this by introducing a robust foundational depth inpainting model for various tasks.

\subsection{Downstream Tasks with Known Partial Depth}
Apart from the traditional depth completion task, many 3D vision tasks involve partially known depth information:
%
(1) 3D scene inpainting focuses on filling in missing parts of 3D spaces, such as removing objects and generating plausible geometry and textures for the inpainted regions. Early works mainly addressed geometry completion~\cite{dai2020sgnn,dai2018scancomplete,kazhdan2006poisson,han2017high,wang2017shape,park2019deepsdf,song2017semantic}, while recent methods focus on jointly inpainting both texuture and geometry~\cite{ye2023gaussian,chen2023gaussianeditor,huang2023point,mirzaei2022laterf,kobayashi2022decomposing,liu2023instance}. Such 3D scenes can be rendered to obtain depth, where unchanged areas serve as known partial depth information.
(2) 3D scene generation involves creating 3D content from inputs such as a single image or text prompt, utilizing strong generative priors and depth estimation. Recent methods~\cite{zhou2025dreamscene360,chung2023luciddreamer,ouyang2023text2immersion,yu2024wonderworld,yu2023wonderjourney,hoellein2023text2room,li2024art3d} change the camera viewpoint, perform RGB inpainting on the warped image, and then estimate depth, aligning it with the warped depth. 
Conversely, our method considers the warped depth as known, formulating the existing two-step process as depth inpainting conditioned at a known scale.
(3) DUST3R~\cite{dust3r_cvpr24} presents a novel 3D reconstruction pipeline that operates on sparse views without any prior knowledge of camera calibration or poses. Recently, several works~\cite{liu2024reconx,seo2024genwarp,sun2024dimensionx} leverage initial point cloud from COLMAP~\cite{colmap} or DUST3R for subsequent tasks such as 3D scene reconstruction~\cite{fan2024instantsplat}, novel view synthesis~\cite{yu2024viewcrafter}, and 4D scene generation~\cite{chu2024dreamscene4d}. However, DUST3R struggles to produce accurate reconstructions in regions without viewpoint overlap. Our method leverages depth data from regions with pixel correspondences to enhance the depth in areas lacking matches.


%

\section{Method}
\begin{figure*}[t]
    \centering
    \includegraphics[width=\linewidth]{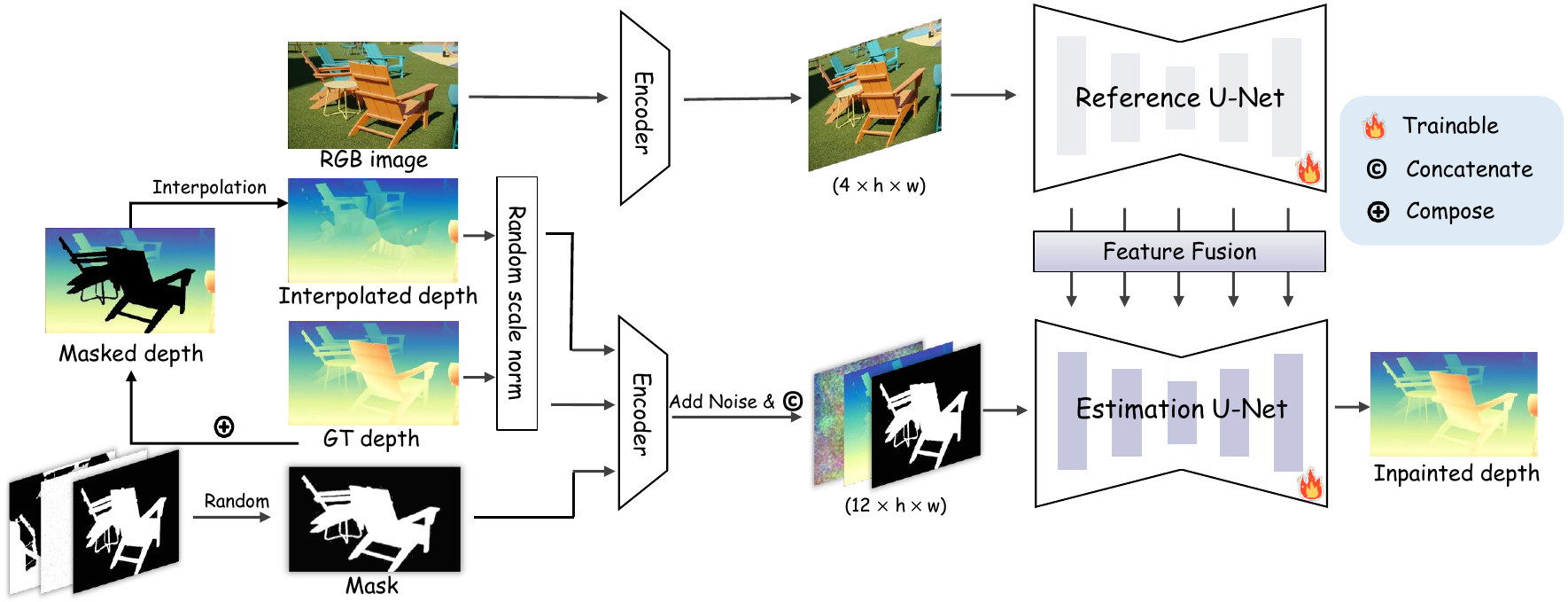}
    \vspace{-20pt}
    \caption{\textbf{The training process of \method.} First, we apply random masking to the ground truth depth to create the masked depth, followed by interpolation. Both the interpolated masked depth and the original depth undergo random scale normalization before being fed into the encoder. The Reference U-Net extracts RGB features, while the Estimation U-Net takes the noisy depth, masked depth, and encoded mask as input. Layer-by-layer feature fusion allows for finer-grained visual guidance, achieving high-quality depth predictions even in large or complex masked regions.}
    \label{fig:pipeline}
    \vspace{-1em}
\end{figure*}
Given an original (incomplete or distorted) depth map $d' \in \mathbb{R}^{1 \times H \times W}$, a binary mask $m \in \mathbb{R}^{1 \times H \times W}$ indicating the target areas for inpainting, and a conditional RGB image $I \in \mathbb{R}^{3 \times H \times W}$, our objective is to use the RGB image to predict a complete depth map $d \in \mathbb{R}^{H \times W}$. This involves preserving the depth values in the unmasked regions while accurately estimating the depth in the masked areas.
This process naturally aligns the estimated depth with the existing geometry, eliminating inconsistencies between the known and target inpainting areas.

To achieve this, we introduce a dual-branch diffusion-based framework for depth inpainting, comprising a Reference U-Net for RGB feature extraction and an Estimation U-Net that takes the original depth and inpainting mask as input.
Instead of the commonly used text conditioning, we utilize cross-attention with the CLIP image encoder to capture rich semantic information.
Layer-by-layer feature fusion through attention between the Reference U-Net and Estimation U-Net enables finer-grained visual guidance.
This design enables \method~to achieve remarkable results, even for large inpainting regions or complex RGB images.
An overview is demonstrated in~\cref{fig:pipeline}.



\subsection{Network Design}
Both branches use Marigold~\cite{marigold} as the base model, fine-tuned from Stable Diffusion V2~\cite{rombach2022high}. 
This design eliminates the need to learn the domain transfer process from RGB to depth, improving training efficiency. 
The detailed architecture of each network is provided below.

\noindent\textbf{Depth encoder and decoder.}
We use a fixed variational autoencoder (VAE) encoder $\mathcal{E}$ to encode both RGB images and their corresponding depth maps into the latent space. 
Since the encoder $\mathcal{E}$ is originally designed for three-channel RGB inputs, single-channel depth maps are replicated across three channels to match the RGB format. 
Notably, as the VAE encoder is intended for non-sparse inputs, we apply nearest neighbor interpolation to densify the sparse depth maps before encoding.
During inference, the denoised depth latent $z^{(d)}_0 \in \mathbb{R}^{4 \times h \times w}$ at step $t=0$ is decoded using the decoder $\mathcal{D}$, and the average of the three channels is used as the predicted depth map.
Unlike Marigold, which estimates relative depth and uses least-squares optimization to obtain metric depth, our depth inpainting aims to directly estimate a metric depth map based on the values and scale of the known depth regions. 
Details regarding depth pre- and post-processing, such as normalization, are provided in~\cref{Training}.

\noindent\textbf{Estimation U-Net.}
The input to the Estimation Denoising U-Net consists of three components: the noisy depth latent $z^{(d)}_t \in \mathbb{R}^{4 \times h \times w}$, the masked depth latent $z^{(d')} \in \mathbb{R}^{4 \times h \times w}$, and the encoded mask $m' \in \mathbb{R}^{4 \times h \times w}$, which are concatenated together. The latent depth representations have 4 channels, resulting from the VAE encoding, and $(h, w)$ are spatial dimensions downsampled by a factor of 8 compared to the original input dimensions $(H, W)$.
Notably, to more accurately preserve mask information, instead of simply downsampling the mask as in traditional image inpainting methods~\cite{ju2024brushnet,zhuang2023task}, we encode the mask $m \in \mathbb{R}^{1 \times H \times W}$ using the VAE encoder $\mathcal{E}$ to obtain $m' \in \mathbb{R}^{4 \times h \times w}$, which effectively retains the sparse and fine-grained information.

During training, the noisy latent depth $z^{(d)}_t$ is obtained by encoding the initial depth map $d$ into the latent space and adding noise at step $t$.
The masked depth latent $z^{(d')}$ is generated by applying random masking to the original ground-truth depth map, followed by nearest-neighbor interpolation in the inpainting regions and encoding via the VAE. Since the VAE of Stable Diffusion~\cite{rombach2022high} excels at reconstructing dense information, this approach better preserves the known depth values at sparse points and complex edge boundaries.

\noindent\textbf{Reference U-Net.}
InFusion~\cite{infusion} feeds a single reference image into the encoder, subsequently concatenating the image latent with noisy depth latent, masked depth latent, and downsampled mask, resulting in a total of 13 channels. 
However, this approach may lose regional depth information or struggle to generate clear depth edges, especially when inpainting large areas or using complex reference images.
Recent studies~\cite{hu2023animateanyone, xu2023magicanimate} demonstrate that an additional U-Net can extract more fine-grained features from the reference image. Inspired by these findings, we adopt a similar architecture. 
We first obtain two feature maps, $\mathbf{f}_1 \in \mathbb{R}^{c \times h \times w}$ and $\mathbf{f}_2 \in \mathbb{R}^{c \times h \times w }$, from the Reference U-Net and Estimation Denoising U-Net, respectively. We then concatenate these feature maps along the width dimension, resulting in $\mathbf{f} \in \mathbb{R}^{c \times h \times 2w}$.
Next, we apply a self-attention operation on the concatenated feature map and extract the first half of the resulting feature map as the output. This allows us to leverage the fine-grained semantic features from each layer of the base model during training.
Furthermore, since the Reference U-Net and Estimation Denoising U-Net share the same architecture and initial weights—both pre-trained on Marigold~\cite{marigold}—the Estimation Denoising U-Net can selectively learn relevant features from the Reference U-Net within the same feature space.

\subsection{Training Protocol} \label{Training}

\noindent\textbf{Depth normalization.}
Our goal is to maintain the original depth information in the known regions, predict the depth for unknown inpainting regions, and avoid geometric inconsistency. 
The final output is an absolute depth map. During inference, since the depth of the inpainting regions is unknown, we cannot determine the minimum and maximum depth values of the entire depth map. 
To simulate this scenario during training, we calculate the minimum and maximum values ($d_\text{min}$ and $d_\text{max}$) for the known depth regions and linearly normalize them to the range of $[-1, 1]$. 
Notably, local minimum and maximum values may not always correspond to the global minimum and maximum, which can lead to overflow during VAE decoding. To address this, we introduce a random compression factor $\beta$ in the range $[0.2, 1.0]$ during normalization to handle these cases more effectively. The depth values are normalized using:
\begin{equation} 
\tilde{d} = \left( \frac{d - d_\text{min}}{d_\text{max} - d_\text{min}} - 0.5 \right) \times 2 \times \beta.
\end{equation} 
This approach adheres to the convention of the Stable Diffusion VAE while also enforcing a canonical, affine-invariant depth representation that is independent of data statistics. 
This ensures that depth values are constrained by the near and far planes, providing stability and reducing the influence of data distribution.
Finally, we apply inverse normalization to the network output to restore the absolute depth scale and achieve depth inpainting.

\noindent\textbf{Masking strategy.}
To maximize coverage across a vast array of downstream tasks, a variety of masking strategies are employed. 
Initially, we randomly select from strokes, circles, squares, or a random combination of these shapes to create the masks. 
Secondly, to enhance the depth completion task—recovering a full depth map from sparse depth data captured by sensors—we implement random point masking, where only 0.1-2\% of points are set as known. 
Lastly, for improved object-level inpainting, we utilize the Grounded-SAM~\cite{ren2024grounded} to annotate training data, subsequently filtering the masks based on their confidence scores. In general, the combined application of multiple masking strategies further enhances the robustness of our method.



\begin{table*}[t]
\caption{\textbf{Quantitative comparison of various methods on different datasets.} Better: AbsRel~$\downarrow$, $\delta_1$~$\uparrow$. The best results are marked in bold, and the second-best underlined. Our method incorporates known depth information, achieving optimal performance across all metrics.
}
\vspace{-1em}
    \centering
    \setlength{\tabcolsep}{1.25mm}{\begin{tabular}{lcccccccccccc}
        \toprule
        & \multicolumn{2}{c}{Training samples} & \multicolumn{2}{c}{NYUv2} & \multicolumn{2}{c}{KITTI} & \multicolumn{2}{c}{ETH3D} & \multicolumn{2}{c}{ScanNet} & \multicolumn{2}{c}{DIODE} \\
        \cmidrule(lr){4-5} \cmidrule(lr){6-7} \cmidrule(lr){8-9} \cmidrule(lr){10-11} \cmidrule(lr){12-13}
        Method & Real &Synthetic & AbsRel$\downarrow$ & $\delta_1$$\uparrow$ & AbsRel$\downarrow$ & $\delta_1$$\uparrow$ & AbsRel$\downarrow$ & $\delta_1$$\uparrow$ & AbsRel$\downarrow$ & $\delta_1$$\uparrow$ & AbsRel$\downarrow$ & $\delta_1$$\uparrow$ \\
        \midrule
        DiverseDepth~\cite{yin2020diversedepth} &320K &--  & 12.1 & 86.8 & 18.8 & 70.2 & 23.0 & 69.9 & 11.1 & 87.6 & 37.2 & 63.8 \\
        MiDaS~\cite{ranftl2020towards} &2M &--  & 10.9 & 88.9 & 24.2 & 62.2 & 18.3 & 75.4 & 13.2 & 87.6 & 33.7 & 70.6 \\
        LeReS~\cite{yin2021learning} &300K &54K  & 9.2 & 91.5 & 14.9 & 78.5 & 17.3 & 77.7 & 9.6 & 90.4 & 27.4 & 77.0 \\
        Omnidata~\cite{eftekhar2021omnidata} &11.9M &310K  & 7.8 & 94.0 & 14.7 & 83.7 & 16.9 & 77.8 & 7.2 & 94.1 & 34.4 & 73.1 \\
        HDN~\cite{zhang2022hierarchical} &300K &--  & 7.2 & 94.6 & 11.2 & 87.2 & 12.1 & 94.2 & 8.0 & 94.2 & 24.2 & 78.3 \\
        DPT~\cite{ranftl2021vision} & 1.2M & 188K & 9.8 & 90.1 & 10.2 & 89.9 & 7.7 & 94.6 & 8.4 & 93.2 & \underline{18.1} & 75.8 \\
        %
        DepthAnthing~\cite{yang2024depth} & 63.5M & -- & \underline{4.4} & 97.6 & 7.6 & 94.7 & 12.5 & 88.5 & 4.2 & 98.1 & 27.4 & 76.1 \\
        DepthAnthingV2~\cite{yang2024depthv2} & 62M & 595K & \underline{4.4} & \underline{98.0} & \underline{7.5} & \underline{94.8} & 13.1 & 86.6 & \underline{4.1} & \underline{98.2} & 27.3 & 76.4 \\
        \midrule
        Marigold~\cite{marigold} & -- & 74k & 5.6 & 96.4 & 9.8 & 91.7 & 6.6 & 95.9 & 6.3 & 95.4 & 30.9 & 77.2 \\
        DepthFM~\cite{gui2024depthfm} & -- & 63K & 6.5 & 95.6 & 8.4 & 93.2 & -- & -- & -- & -- & 22.4 & \underline{79.8}\\
        GeoWizard~\cite{geowizard} & -- & 278K & 5.2 & 96.5 & 9.6 & 92.3 & \underline{6.4} & \underline{96.3} & 6.1 & 95.4 & 29.5 & 79.5 \\
        %
        %
        \midrule
        Ours & -- & 74k & \textbf{2.5} & \textbf{98.8} & \textbf{7.2} & \textbf{95.3} & \textbf{3.1} & \textbf{97.9} & \textbf{2.3} & \textbf{98.5} & \textbf{17.6} & \textbf{85.6} \\
        \bottomrule
    \end{tabular}}
    \vspace{-1em}
    \label{tab:results}
\end{table*}
\begin{figure*}[t]
    \centering
    \includegraphics[width=\linewidth]{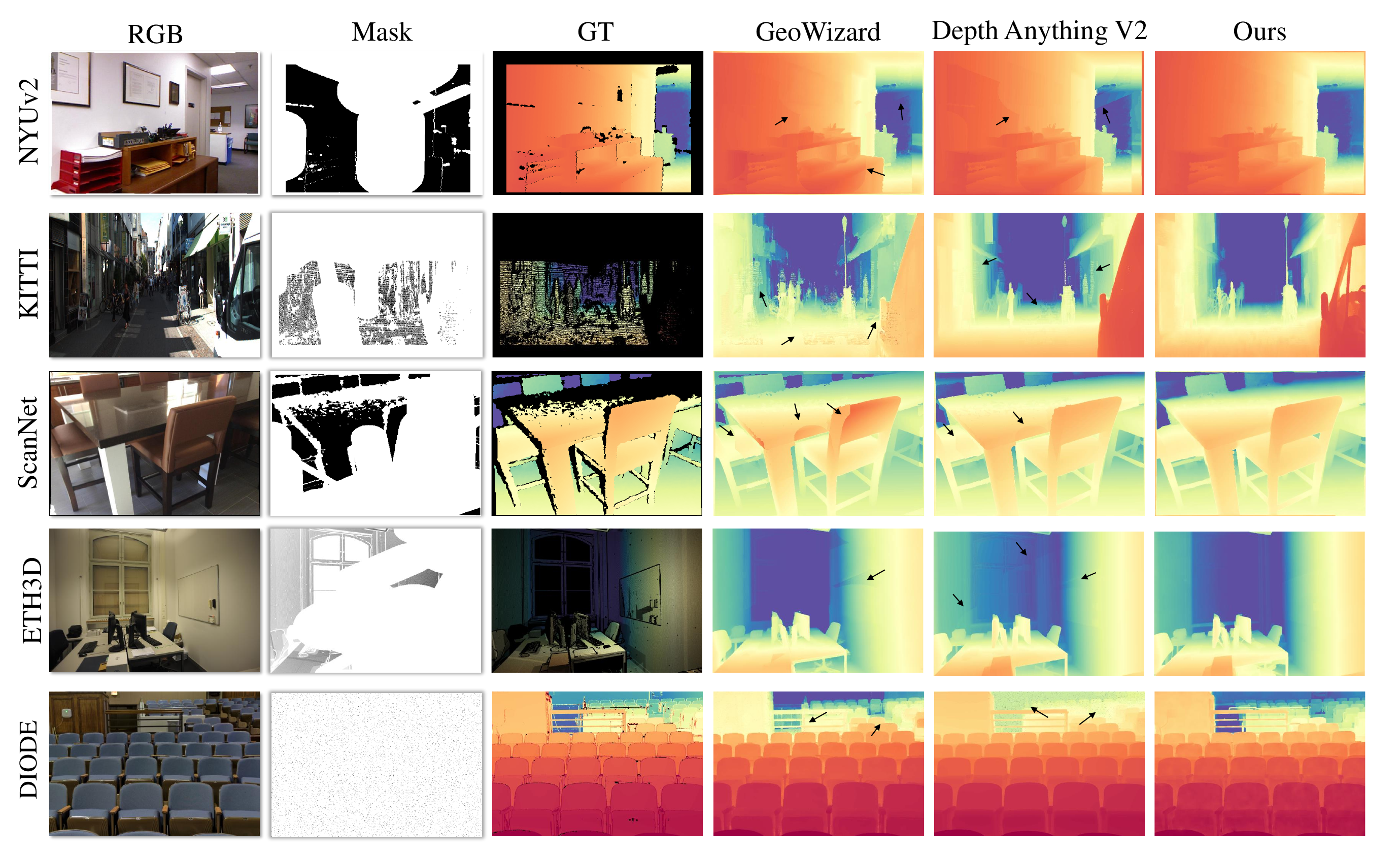}
    \vspace{-2em}
    \caption{\textbf{Qualitative comparison of various methods on different datasets.} In the second column, black represents the known regions, while white indicates the predicted areas. 
    Notably, to emphasize the contrast, we reattach the known ground truth depth to the corresponding positions in the right-side visualizations of the depth maps. Other methods exhibit significant geometric inconsistency.
    }
    \label{fig:compare}
    \vspace{-1em}
\end{figure*}

\begin{figure*}[t]
\vspace{-1em}
    \centering
    \vspace{-1em}
    \includegraphics[width=\linewidth]{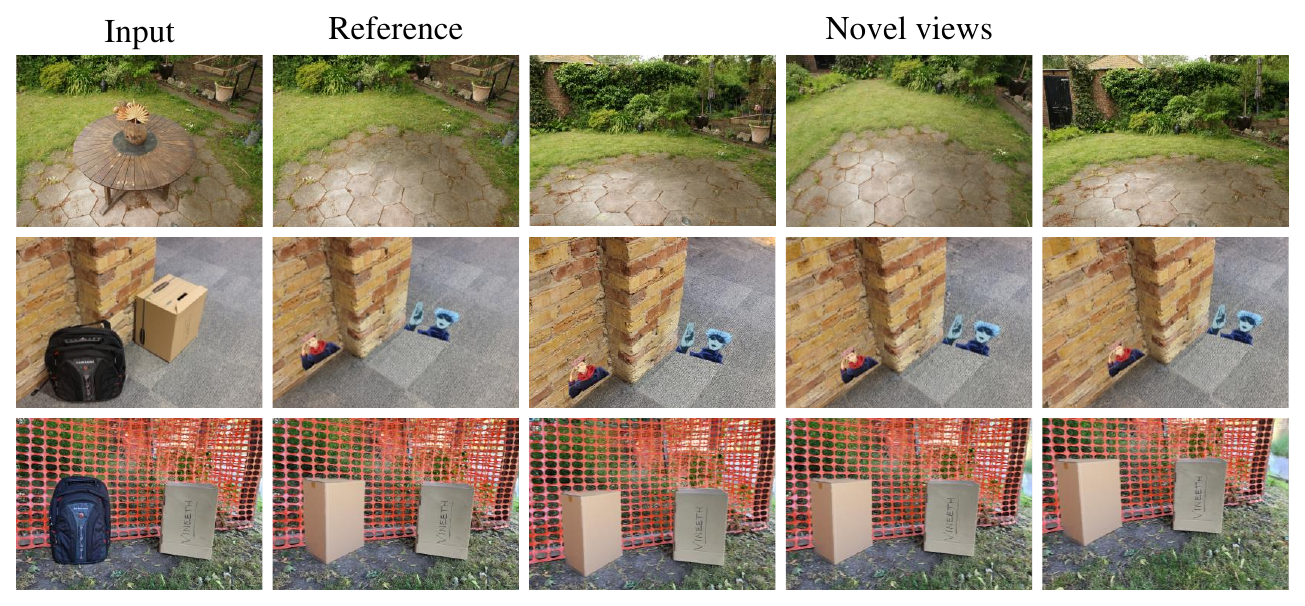}
    \vspace{-2em}
    \caption{\textbf{Visualization of gaussian inpainting.} By projecting depth directly into three-dimensional space as initial points, natural 3D consistency is maintained, enabling texture editing and object addition. Please zoom in to view more details.}
    \label{fig:inpainting}
    \vspace{-1em}
\end{figure*}

\section{Experiments}
\subsection{Implementation Details and Datasets}
\noindent\textbf{Training details.}
For training \method, we utilize two synthetic datasets covering indoor and outdoor scenes. The first dataset, Hypersim~\cite{roberts:2021}, is a photorealistic dataset with 461 indoor scenes, from which we select 365 scenes, totaling approximately 54K training samples after filtering out incomplete samples.
The second dataset, Virtual KITTI~\cite{cabon2020vkitti2}, is a synthetic street-scene dataset comprising five scenes with varying conditions, such as weather and camera perspectives. We employ 4 of these scenes, amounting to approximately 20K training samples.
We initialize the Reference U-Net and Estimation U-Net with pre-trained weights from the Marigold~\cite{marigold}. 
We also test other pre-trained weights, and the results are presented in Supplementary Materials. 
The training process spans 200 epochs, starting with an initial learning rate of 1e-3. This learning rate is scheduled to decay after every 50 epochs. We only use random flips as data augmentation. 
Utilizing eight A100-80G GPUs, the training process is completed within two days.

\noindent\textbf{Evaluation datasets.}
We assess our performance across five zero-shot benchmarks, including NYUv2~\cite{Silberman:ECCV12}, KITTI~\cite{geiger2013vision}, ETH3D~\cite{schops2017multi}, ScanNet~\cite{dai2017scannet}, and DIODE~\cite{vasiljevic2019diode}. During inference, we randomly select strokes, circles, squares, or combinations of the three across the entire image to represent unknown regions. 
In addition, we include another type of mask, where we randomly set only 0.5\% to 1\% of sparse pixels across the entire image as known depth.

\begin{figure*}[t]
    \centering
    \includegraphics[width=\linewidth]{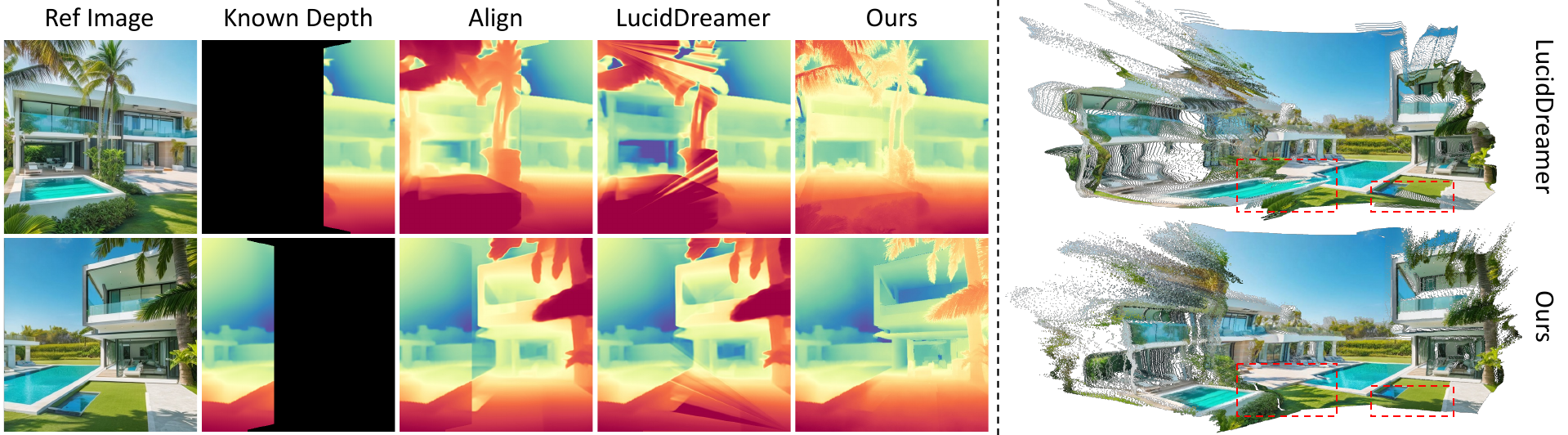}
    \vspace{-1.5em}
    \caption{\textbf{Visualization of 3d scene generation.} \textbf{Left}: Depth comparison. "Align" represents the least-square method and shows clear geometric inconsistencies at boundaries. While LucidDreamer reduces these inconsistencies, it compromises the accuracy of the newly estimated depth. 
    In contrast, our model produces consistent and accurate depth. \textbf{Right}: The improved depth estimation from our model leads to superior 3D scene generation results.
}
    \vspace{-1em}
    \label{fig:lucid}
\end{figure*}

\begin{figure*}[t]
    \centering
    \includegraphics[width=\linewidth]{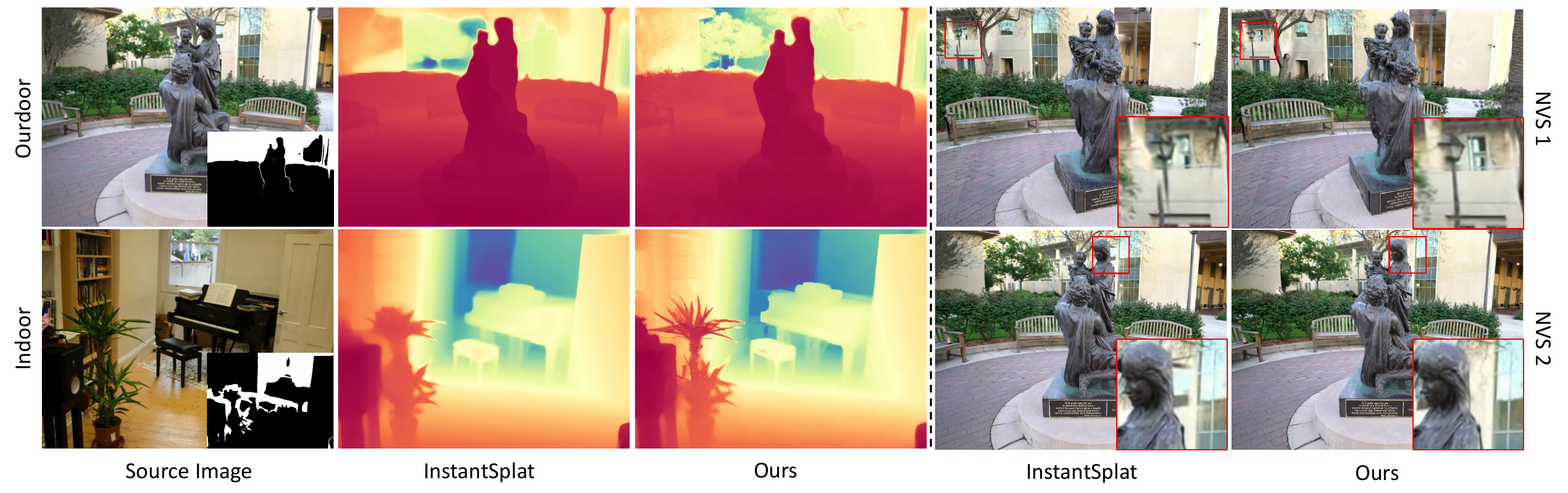}
    \vspace{-2em}
    \caption{\textbf{Visualization of sparse-view reconstruction with DUST3R.} \textbf{Left}: Compared to InstantSplat~\cite{fan2024instantsplat}, which directly uses point cloud from DUST3R as initialization, our method produces sharper and clearer depth in non-matching regions. \textbf{Right}: Using our method for improved initialization results in higher-quality Gaussian splatting rendering. Please zoom in for details.}
    \label{fig:duster}
\end{figure*}

\subsection{Comparisons}
Given an image with ground truth depth, we first mask some regions of the GT depth to simulate missing data and perform depth inpainting using our model. For other methods, we utilize their pretrained models to estimate the depth map from the corresponding RGB image. Metric calculations are restricted to the masked regions to assess the inpainting performance while excluding the known regions from the evaluation. 

\noindent\textbf{Baselines and metrics.}
We compare \method~with both discriminative and generative methods. The discriminative methods include DiverseDepth~\cite{yin2020diversedepth}, MiDaS~\cite{ranftl2020towards}, LeReS~\cite{yin2021learning}, Omnidata~\cite{eftekhar2021omnidata}, HDN~\cite{zhang2022hierarchical}, DPT~\cite{ranftl2021vision}, DepthAnything~\cite{yang2024depth}, and DepthAnythingV2~\cite{yang2024depthv2}. 
The generative methods include Marigold~\cite{marigold}, DepthFM~\cite{gui2024depthfm}, and GeoWizard~\cite{geowizard}.
Following prior research~\cite{yin2023metric3d}, we evaluate the performance of depth estimation methods using absolute relative error (AbsRel) and accuracy within a threshold of $\delta^{1} = 1.25$.

\noindent\textbf{Qualitative comparison.}
We present the quantitative evaluations in~\cref{tab:results}.
Notably, unlike discriminative methods such as DepthAnything V2~\cite{yang2024depthv2}, which is trained on a dataset of $63$M samples, our approach relies solely on $74$K synthetic data.
Compared to other generative methods, our approach not only inherits the rich priors of the pre-trained diffusion model but also leverages its ability to incorporate local priors for conditionally generating coherent images.
This enables us to effectively utilize partial known depth and align with the known scale, achieving optimal performance across all datasets and metrics.

\noindent\textbf{Quantitative comparison.}
As shown in~\cref{fig:compare}, we set the black areas in the mask as known depth and the white areas as regions for prediction. 
Notably, for ease of comparison, we replace the known regions with ground truth in the depth maps shown in the three rightmost columns. 
The depth estimated by our method exhibits better geometric consistency, while other monocular methods, even after least-squares optimization, still show noticeable disjunctions with the known regions. Please zoom in to observe the details.

\subsection{Applications}
In this section, we provide a detailed overview for utilizing \method~in downstream tasks and present the corresponding results.

\noindent\textbf{3D gaussian inpainting.}
Initially, we employ Gaussian Grouping~\cite{gaussian_grouping} to segment and remove partial Gaussians.
The SDXL Inpainting model~\cite{podell2023sdxl} is then applied to the rendered image at the reference view. 
The inpainted RGB image subsequently serves as guidance to complete the depth information for that reference view.
The points are then unprojected into three-dimensional space for optimal initialization. 
As shown in~\cref{fig:inpainting}, due to the geometric consistency between the inpainted and original Gaussians and the alignment between pixels and inpainted Gaussians, simple edits to the inpainted image enable texture modification and object insertion in the inpainted regions.

\begin{table*}[t]
\caption{\textbf{Quantitative comparison of depth completion.}"Ours*" represents the zero-shot capability of our model, while "Ours" represents its performance after fine-tuning.}
    \centering
    \setlength{\tabcolsep}{1.0mm}{\begin{tabular}{c|ccccccccc}
        \hline
         & NLSPN~\cite{park2020non} & DSN~\cite{de2021deep} & Struct-MDC~\cite{jeon2022struct} &ACMNet~\cite{zhao2021adaptive}& CFormer~\cite{zhang2023completionformer} & BP-Net~\cite{tang2024bilateral} & LRRU~\cite{wang2023lrru} &Ours* & Ours \\ \hline \textbf{RMSE}
        & 0.092 & 0.102 & 0.245  & 0.105 & 0.090 & \textbf{0.089} & 0.091 & 0.104 & 0.090\\ \hline
    \end{tabular}}
    \label{tab:completion}
    \vspace{-1em}
\end{table*}

\begin{table*}[t]
\caption{\textbf{Analysis of known depth ratios}. We assess our model's performance at known depths of 2\%, 5\%, 10\%, 30\%, and 50\% ratio.}
\label{tab:ratio}
    \centering
    \setlength{\tabcolsep}{1.8mm}{\begin{tabular}{ccccccccccccc}
        \hline
         &\multicolumn{2}{c}{2\%}  &\multicolumn{2}{c}{ 5\% }& \multicolumn{2}{c}{10\%} & \multicolumn{2}{c}{30\%}  &\multicolumn{2}{c}{ 50\%}
         \\
         \cmidrule(lr){2-3} \cmidrule(lr){4-5} \cmidrule(lr){6-7} \cmidrule(lr){8-9} \cmidrule(lr){10-11} \cmidrule(lr){12-13}
          Dataset  & AbsRel$\downarrow$ & $\delta_1$$\uparrow$ & AbsRel$\downarrow$ & $\delta_1$$\uparrow$ & AbsRel$\downarrow$ & $\delta_1$$\uparrow$ & AbsRel$\downarrow$ & $\delta_1$$\uparrow$ & AbsRel$\downarrow$ & $\delta_1$$\uparrow$ \\
        \midrule
        \textbf{NYUv2}
        & 3.3 & 98.2 & 3.0 & 98.3 & 2.8  & 98.4  & 2.5 & 98.8  & 2.2 & 98.8\
        \\
        \textbf{ETH3D}
        & 3.1 & 97.4 & 2.9 & 98.0 & 2.7  & 98.3  & 2.3 & 98.5  & 2.0 & 98.6\
        \\
         \hline
    \end{tabular}}
    \vspace{-1em}
\end{table*}

\noindent\textbf{Text to 3D scene generation.}
Recent methods~\cite{zhou2025dreamscene360,chung2023luciddreamer,ouyang2023text2immersion} begin by projecting single-view depth estimation onto a 3D scene to create an initial point cloud from a given viewpoint. 
The camera is then rotated to compute the warped image and warped depth. 
Following inpainting on the warped RGB image, monocular depth estimation is applied, and the estimated depth is aligned with the previously warped depth. 
The aligned depth data is then unprojected back into the original point cloud. This process repeats with changes in viewpoint. 
However, as shown in~\cref{fig:lucid}, using LucidDreamer~\cite{chung2023luciddreamer} as an example, this approach suffers from geometric inconsistency during depth alignment across different scales, adversely impacting depth accuracy in the inpainted regions.
In contrast, our model can directly take the inpainted image and warped depth as input, producing geometrically consistent depth maps without requiring alignment.

\noindent\textbf{Reconstruction with DUST3R.}
DUST3R~\cite{dust3r_cvpr24} introduces the new 3D reconstruction pipeline from sparse views without prior information about camera calibration or poses. 
This method can provide dense pixel-to-pixel and pixel-to-point correspondences, making it applicable to various 3D vision tasks, including depth prediction. 
However, we observe that DUST3R delivers high-quality depth predictions primarily at points with pixel correspondences while struggling to generate clear depth edges for points without correspondence across different views.

To overcome this limitation, we introduce a refinement framework that improves depth estimation for regions with weak or absent correspondences. 
Our approach begins by generating a mask for pixels without matches from any source images. 
These non-matching regions are then refined through our proposed \method.
Specifically, we employ a variational autoencoder (VAE) to encode the initial depth estimates from DUST3R into the latent space, adding noise to produce noisy latents. 
Depths at matched points are encoded as masked depth latents.
Both the noisy and masked latent representations, along with their respective masks, are fed into our model to generate refined depth maps with enhanced accuracy and spatial consistency.
We evaluate our method on InstanSplat~\cite{fan2024instantsplat}, a sparse-view, SfM-free Gaussian splatting method that uses the predicted point cloud from DUST3R for novel view synthesis.
By unprojecting our enhanced depth maps into 3D space, we replace DUST3R's original point cloud with our refined data as input for InstanSplat.
As shown in~\cref{fig:duster}, our approach effectively sharpens initial depth from DUST3R, substantially improving Gaussian splatting rendering quality. We also provide quantitative comparisons in the supplementary materials.

\noindent\textbf{Sensor depth completion.}
Sensor depth completion is an important task related to depth estimation, with widespread applications in robotics and autonomous navigation. Due to hardware limitations of depth sensors, only a partial depth image is available, making it necessary to fill in the missing depth values while preserving the known depth values.
We obtain the corresponding masks based on the coordinates of the sparse depth and perform comparisons on the NYU Depth v2~\cite{Silberman:ECCV12}. Following CompletionFormer~\cite{zhang2023completionformer}, during evaluation, the original frames of size $640 \times 480$ are downsampled by half using bilinear interpolation, then center-cropped to $304 \times 228$, with only $500$ ground-truth pixels available. As shown in~\cref{tab:completion}, our model differs from traditional depth completion methods, which are typically trained and tested on fixed datasets, leading to limited generalization capabilities. In contrast, our approach exhibits better generalization without requiring complex designs, and with fine-tuning of only $10,000$ steps, it achieves performance comparable to state-of-the-art methods, demonstrating its potential as a foundational model for depth completion tasks.
Additionally, one major limitation is the need for downsampling the mask in the latent space, where the VAE of Stable Diffusion 2.1 struggles to finely reconstruct extremely sparse data. Further analysis is provided in the supplementary materials.
\subsection{Analysis of Known Depth Ratios.} \label{ablation}
To investigate the dependency of our model's performance on known depth ratios, we randomly select varying proportions of pixels with ground truth depth information to serve as known conditions for inference. As shown in~\cref{tab:ratio}, our model demonstrates outstanding performance even with very limited known conditions.
We will provide more analysis on the known depth in the supplementary materials.

\section{Conclusion and Future Work}

\noindent\textbf{Conclusion.}
In this work, we introduced DepthLab, a robust depth inpainting framework designed to handle complex scenarios by leveraging RGB images and known depth as conditional inputs. Our approach maintains geometric consistency, ensuring that estimated depth aligns seamlessly with existing structures. By leveraging the priors in pre-trained diffusion models, DepthLab demonstrates significant generalization across various depth inpainting tasks. This is evidenced by its superior performance on multiple benchmarks and applications, including 3D scene generation, Gaussian inpainting, LiDAR depth completion, and sparse-view reconstruction. Our experiments showcase the robustness of our method, highlighting its ability to improve both depth accuracy and downstream 3D scene quality. Moving forward, we envision DepthLab serving as a foundational model for a wider range of tasks.

\noindent\textbf{Future work.}
We expect three promising directions for future research: (1) employing consistency model~\cite{song2023consistency} or flow-based approaches~\cite{lipman2022flow, wang2024rectified} to accelerate sampling speed; (2) further fine-tuning existing image VAEs to more effectively encode sparse information, such as sparse masks and depth; and (3) extending this inpainting technique to normal estimation models, which could facilitate more versatile editing of different 3D assets.
\section*{Acknowledgements}
This paper is partially supported by the National Key R\&D Program of China No.~2022ZD0161000 and the General Research Fund of Hong Kong No.~17208825.
\bibliographystyle{ieeenat_fullname}
\bibliography{main}

\end{document}